\documentclass[10pt,conference]{IEEEtran}
\IEEEoverridecommandlockouts

\usepackage{cite}
\usepackage{amsmath,amssymb,amsfonts}
\usepackage{algorithmic}
\usepackage{graphicx}
\usepackage{textcomp}
\usepackage{xcolor}
\usepackage{subcaption}
\usepackage{hyperref}
\usepackage[inline]{enumitem}
\usepackage{cleveref}
\crefname{figure}{Fig.}{Figs.}
\crefname{table}{Table}{Tables}
\usepackage{lipsum}
\usepackage{booktabs}
\usepackage{array}
\usepackage{acronym}
\usepackage{todonotes}
\usepackage[T1]{fontenc}
\usepackage[skins,breakable]{tcolorbox}
\usepackage{listings}
\usepackage{balance}
\usepackage{microtype}
\usepackage[subtle]{savetrees}
\usepackage{pifont}
\usepackage{siunitx}
\usepackage{multirow}

\tcbuselibrary{listings}

\newcounter{promptlisting}

\newtcblisting{promptlisting}[2]{
    enhanced,
    breakable,
    listing only,
    colback=gray!5,
    colframe=gray!40,
    boxrule=0.5pt,
    arc=2pt,
    left=6pt,
    right=6pt,
    top=6pt,
    bottom=6pt,
    title={#1},
    fonttitle=\bfseries,
    listing options={
        basicstyle=\ttfamily\scriptsize,
        breaklines=true,
        breakatwhitespace=true,
        columns=fullflexible,
        keepspaces=true,
        breakindent=0pt,
        breakautoindent=false,
        showstringspaces=false,
        upquote=true
    }
}

\newcounter{promptbox}
\renewcommand{\thepromptbox}{\arabic{promptbox}}

\newenvironment{promptbox}[2]
{%
    \refstepcounter{promptbox}%
    \label{#2}%
    \begin{tcolorbox}[
        enhanced,
        breakable,
        colback=gray!10,
        colframe=gray!300,
        colbacktitle=gray!300,
        boxrule=0.5pt,
        arc=2pt,
        left=6pt,
        right=6pt,
        top=6pt,
        bottom=6pt,
        title={Listing~\thepromptbox: #1},
        fonttitle=\bfseries,
    ]%
     \footnotesize
}
{%
    \end{tcolorbox}%
}

\lstdefinestyle{promptstyle}{
    backgroundcolor=\color{gray!10},
    basicstyle=\ttfamily\scriptsize,
    breaklines=true,
    frame=single,
    columns=fullflexible,
    keepspaces=true,
    showstringspaces=false,
    breakatwhitespace=true
}

\acrodef{ai}[AI]{Artificial Intelligence}
\acrodef{cot}[CoT]{Chain of Thoughs}
\acrodef{dnn}[DNN]{Deep Neural Network}
\acrodefplural{dnn}[DNNs]{Deep Neural Networks}
\acrodef{llm}[LLM]{Large Language Model}
\acrodefplural{llm}[LLMs]{Large Language Models}
\acrodef{ml}[ML]{Machine Learning}
\acrodef{rag}[RAG]{Retrival Augmented Generation}
\acrodef{rlhf}[RLHF]{Reinforcement Learning from Human Feedback}
\acrodef{rq}[RQ]{Research Question}
\acrodefplural{rq}[RQs]{Research Questions}
\acrodef{sas}[SAS]{Self-Adaptive Systems}
\acrodef{genai}[GenAI]{Generative Artificial Intelligence}
\acrodef{iot}[IoT]{Internet of Things}
\acrodef{pomdp}[POMDP]{Partially Observable Markov Decision Process}

\crefname{lstlisting}{listing}{listings}
\Crefname{lstlisting}{Listing}{Listings}

\def\cameraready{}  

\begin{document}

\title{CVE2AP: Automated Generation of PDDL-Encoded Attack Paths via Large Language Models 
\thanks{This work was supported by funding from the Helmholtz Association (HGF), topic 46.23 Engineering Secure Systems (KASTEL) and from the HGF pilot program Core Informatics at KIT (KiKIT)}
}
\ifdefined\cameraready
    \author{
        \IEEEauthorblockN{Lin Cui, Vincenzo Scotti, Raffaela Mirandola}
        \IEEEauthorblockA{
            \textsuperscript{\textdagger}\textit{KASTEL}, \textit{Karlsruhe Institute of Technology}, Karlsruhe, Germany \\
           \texttt{lin.cui@kit.edu}, \texttt{vincenzo.scotti@kit.edu}, \texttt{raffaela.mirandola@kit.edu}}
    }
\else
    \author{
        Anonymous Author(s)
    }
\fi

\maketitle

\bstctlcite{IEEEexample:BSTcontrol}

\begin{abstract}
Attack Path (AP) modeling is fundamental to cybersecurity analysis, where the Planning Domain Definition Language (PDDL) has been widely adopted to encode APs into formal and machine-verifiable representations for automated reasoning about vulnerability exploitation, attack progression, and their potential impacts.
However, existing AP modeling approaches largely rely on expert-driven manual construction, limiting their scalability and ability to keep pace with rapidly evolving cyber threats.
Large language models (LLMs) are promising candidates, as their extensive pre-trained knowledge and reasoning capabilities enable them to interpret and transform threat intelligence into formal representations.
In this paper, we propose \textbf{CVE2AP}, an LLM-based approach for automatically generating PDDL-encoded attack paths from natural language CVE (Common Vulnerability Exposure) descriptions. CVE2AP leverages structured prompting and incorporates an error-feedback mechanism that iteratively refines the generated paths using planner-reported syntactic and solvability errors.
We conduct a systematic empirical evaluation across multiple LLMs and generation configurations, assessing generation quality across syntactic, solvability and semantic dimensions, together with token consumption and generation time. The results demonstrate that CVE2AP effectively generates high-quality PDDL-encoded attack paths, achieving up to 86.9\% syntax correctness, 78.6\% solvability, and 93.1\% semantic correctness under LLM-as-expert evaluation, while \texttt{GPT-5.5} offers the best quality-cost trade-off and error feedback yields the most consistent quality improvement.
\end{abstract}
\acresetall{}

\begin{IEEEkeywords}
Automated Generation, Attack Path, PDDL, Semantic Quality
\end{IEEEkeywords}






\section{Introduction}
\label{sec:introduction}
Formal and machine-verifiable Attack Path (AP) models provide foundation for cybersecurity analysis by enabling automated reasoning about vulnerability exploitation, attack progression, and their potential impacts~\cite{sheyner2002automated}. To support such reasoning, the Planning Domain Definition Language (PDDL)~\cite{aeronautiques1998pddl}, originally designed for AI planning, has been adapted for representing attack paths~\cite{roberts2011personalized, obes2013attack, tiwary2017pddlassistant, falco2018master, bashir2024resiliency, inayat2025employing, bezawada2019agbuilder}. By explicitly specifying the required preconditions and effects of every action, PDDL naturally models the causal progression of an attack from the initial vulnerability exposure to the resulting security impact. 
However, driven by the continuous disclosure of new vulnerabilities, the emergence of sophisticated attack techniques, and the rapid accumulation of cyber threat intelligence, the cyber threat landscape is evolving at an unprecedented pace. This situation require that attack paths must be continuously and rapidly updated to reflect newly feasible attack strategies and changing attack conditions. As a result, manually constructing and maintaining attack paths can not keep pace with this rapid evolution. It is of urgent need to automate the generation of PDDL-encoded attack paths.

Existing PDDL-based attack path modeling approaches~\cite{roberts2011personalized, obes2013attack, falco2018master,bashir2024resiliency, inayat2025employing} largely rely on domain experts to manually construct PDDL-encoded attack paths.  
Creating such planner-ready representations requires translating heterogeneous incomplete, and often implicit cyber threat intelligence into formal PDDL models, demanding expertise in both cybersecurity and AI planning.
Consequently, developing and maintaining such models is labor-intensive and difficult to scale. 
In parallel, recent studies have demonstrated the effectiveness of LLMs in automatically generating general-purpose PDDL domains from natural language inputs~\cite{guan2023leveraging, smirnov2024generating, oswald2024large, huang2025limit, huang2025planning}. While these methods substantially reduce the manual effort of planning domain construction, their applicability of PDDL-encoded attack path generation remains largely unexplored. Unlike general planning domains, PDDL-encoded attack paths must satisfy not only syntax correctness and planning solvability, but also cybersecurity-specific semantic requirements including threat model compliance, exploitation mechanism fidelity, attack stage completeness.
These observations raise a fundamental research challenge: \emph{how can PDDL-encoded attack paths be automatically generated while satisfying syntactic correctness, planning solvability, and cybersecurity-specific semantic requirements?}

In this paper, we address this challenge by proposing \textbf{CVE2AP}, an LLM-based approach for automatically generating PDDL-encoded attack paths from CVE descriptions.
CVE2AP combines structured prompting with error-feedback mechanism that iteratively refines generated results using planner-reported syntax and solvability errors. 
We further conduct a systematic empirical study to evaluate the effectiveness as well as token and time costs of CVE2AP across mutiple LLMs and generation configurations.
In summary, the main contributions of this paper are:
\begin{itemize}[leftmargin=1em]
     \item We propose CVE2AP, an LLM-based approach for automatically generating PDDL-encoded attack paths from natural language CVE descriptions while satisfying both planning surface-level requirements and cybersecurity-specific semantic requirements.
     \item We investigate generation performance across multiple LLMs and generation configurations, evaluating generation quality using the S3Eval framework~\cite{cui2026s3eval} from syntactic, solvability, and semantic perspectives, together with token consumption and generation time.
     
\end{itemize}

The remainder of this paper is organized as follows. Section~\ref{sec:background} introduces the necessary background. Section~\ref{sec:methodology} presents the proposed CVE2AP approach. Section~\ref{sec:evaluation} describes the experimental setup and analyzes the results. Section~\ref{sec:related-work} discusses related work. Section~\ref{sec:conclusion} concludes the paper and outlines future research directions.

\section{Background}
\label{sec:background}
This section presents the problem statement (Section~\ref{sec:problem-statement}) and the formal representation used to model attack paths (Section~\ref{sec:pddl-for-ap}).

\subsection{Problem Statement}
\label{sec:problem-statement}
Formal attack paths provide a machine-processable representation of vulnerability exploitation, attack progression, and resulting security impacts, supporting automated cybersecurity analysis. 
CVE descriptions contain rich information about vulnerabilities, their exploitation conditions, and security impacts, making them a natural source for deriving attack paths.
For example, CVE-2023-6378 describes a serialization vulnerability in the Logback receiver component of Logback 1.4.11 that allows an attacker to mount a denial-of-service attack by sending poisoned data. 

\begin{tcolorbox}[
    colback=gray!5,
    colframe=gray!40,
    boxrule=0.4pt,
    arc=1pt,
    left=6pt,
    right=6pt,
    top=5pt,
    bottom=5pt
]
\textbf{CVE-2023-6378.}
A serialization vulnerability in logback receiver component part of logback version 1.4.11 allows an attacker to mount a Denial-Of-Service attack by sending poisoned data.
\end{tcolorbox}

This description provides rich information relevant to exploting the vulnerability, such as the vulnerability characteristics, affected component and version, exploitation mechanism, and intended security impact.
However, this information is expressed in natural language and leaves the concrete attack actions and their dependencies implicit. 
Generating an attack path from a CVE description therefore requires making this implicit attack logic explicit as a feasible sequence of attack actions.

We represent the resulting attack path in PDDL to obtain a formal, planner-ready representation that supports automated verification and reasoning.
Formally, given a natural-language CVE description $C$, the task is to generate a PDDL-encoded attack path $P=f(C)$ that connects the vulnerability to its resulting security impact through a feasible sequence of attack actions.
The generated representation must satisfy both formal planning and cybersecurity-specific requirements including \emph{syntactic validity}, \emph{planning solvability}, and \emph{cybersecurity semantic fidelity}.

\subsection{PDDL for Attack Path Modeling}
\label{sec:pddl-for-ap}
The PDDL~\cite{aeronautiques1998pddl} is a standard formal language for automated planning.
A PDDL \emph{domain} defines
\emph{action} schemas, each with typed parameters, preconditions that must hold for the action to be applicable, and effects that describing the resulting world state upon the action's execution. A PDDL \emph{problem} grounds the domain for a concrete scenario by specifying initial states and goals. A domain-independent planner then searches for a plan, i.e., a sequence of grounded actions that transforms the initial state into one satisfying the goal.

In the context of cybersecurity, PDDL has been adapted to model attack paths representing the progression of an attack against a vulnerable system~\cite{roberts2011personalized, tiwary2017pddlassistant, bezawada2019agbuilder}. Each action indicates an operation by attacker, target system, user or infrastructure, with preconditions encoding required capabilities or conditions, and effects capturing the resulting states upon execution. The attack path typically follows stages standardized by the \emph{Cyber Kill Chain}~\cite{hutchins2011intelligence}: starting with Reconnaissance, where system exposures and vulnerabilities are identified, followed by Weaponization and Delivery, during which malicious scripts, files, or payloads to exploit the known vulnerability are crafted and transmitted to the target. This is followed by Exploitation that achieves the exploitation goal with a final impact. The resulting impact can be categorized by the established threat classification framework such as \emph{STRIDE}~\cite{shostack2014threat}: Spoofing, Tampering, Repudiation, Information Disclosure, Denial of Service, and Elevation of Privilege.

Figure~\ref{fig:pddl-example} presents a simplified PDDL-encoded attack path. The first action corresponds to the Reconnaissance stage, identifying that the target system is exposed and accepts external requests. The second action models the Weaponization stage, during which the attacker crafts a malicious payload. 
Upon execution of step-wise actions, the final steps achieves Denial of service by the successful Exploitation.



\begin{figure}[htbp]
\centering
\begin{lstlisting}[style=promptstyle]
;Step 0: Reconnaissance, identify exposed system
(:action system-exposes-attack-surface
  :parameters (?System - system ?Software - software)
  :precondition (and
    (runs-vulnerable-software ?System ?Software)
    (<= (version ?Software) 1001020001) ;version <= 1.1.20.1 
    (accepts-external-requests ?System))
  :effect (and
    (exposes-attack-surface ?System)
    (target-accepts-requests ?System)))
;Step 1: Weaponization
(:action attacker-crafts-malicious-payload
  :parameters (?System - system ?Payload - payload)
  :precondition (exposes-attack-surface ?System)
  :effect (crafted-payload ?Payload))
; ... Omitted steps
;Step N: Successful Exploitation (STRIDE impact)
(:action attacker-achieves-denial-of-service
  :parameters (?System - system)
  :precondition (remote-code-execution ?System)
  :effect (denial-of-service ?System))
\end{lstlisting}
\caption{Simplified example of a PDDL-encoded attack path.}
\label{fig:pddl-example}
\end{figure}

\section{Methodology}
\label{sec:methodology}
\begin{figure*}[htbp!]
    \centering
    \includegraphics[width=\linewidth]{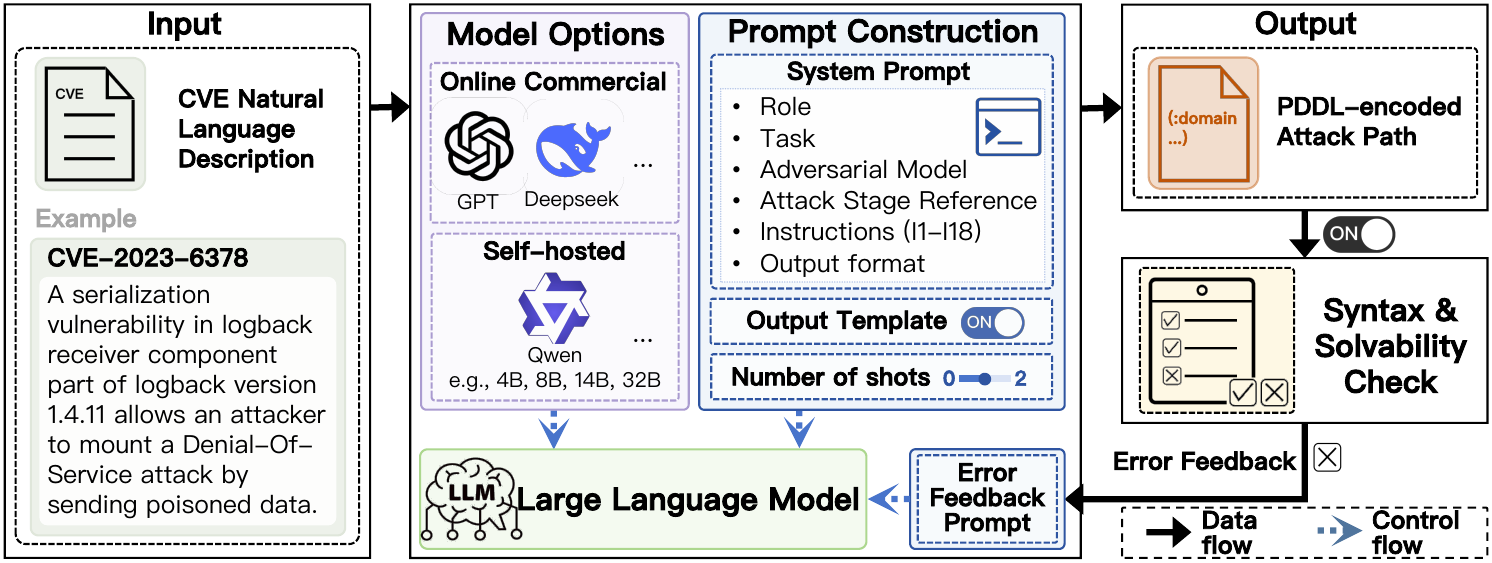}
    \caption{The overview of the CVE2AP.}
    \label{fig:cve2ap-overview}
\end{figure*}

This section presents CVE2AP, an LLM-based approach for automatically generating PDDL-encoded attack paths from CVE natural language descriptions.
Figure~\ref{fig:cve2ap-overview} provides an overview.
Given a natural-language CVE description, CVE2AP generates a PDDL-encoded attack path as output. The approach is model-agnostic and can be instantiated with any LLM that supports either a chat-based or completion-based interface. The generation process is governed by two complementary components: (i) \emph{prompt construction},
which provides the LLM with task-specific knowledge and assembles the generation prompt from a system prompt, optional few-shot examples, and an optional output template; and (ii) an optional \emph{error feedback} mechanism, which automatically validates the generated PDDL using a domain-independent planner and iteratively refines invalid or unsolvable outputs based on the  planner's feedback.

\subsection{Model Options}
CVE2AP is model-agnostic and can be instantiated with any LLM that supports a chat-based or completion-based interface. To cover diverse deployment scenarios and enable analysis across different model scales, we considers two aspects of model instantiation: (i)~\emph{deployment mode}, including online commercial models accessed through remote APIs and self-hosted models deployed locally, and (ii)~\emph{model scale}, 
considering multiple model sizes within the same model family to isolate, as far as possible, the effect of model scale on attack-path generation quality.
The specific model instances used in our experiments are detailed in Section~\ref{sec:ex-setup}.

\subsection{Prompt Construction}
\label{sec:prompt-construction}
CVE2AP supports both structured chat-based interaction, using system--user--assistant messages for instruction-tuned models, and plain completion-based interaction for base models. 
The appropriate interaction format is automatically selected according to the chosen model. 
To ensure consistent generation behavior across models, the core prompt content, including the system prompt, task instructions, and few-shot examples, is defined through a common base template and rendered into the corresponding interaction format. The prompt comprises the following components.

\textbf{System Prompt.} The system prompt defines the LLM's role as an expert in both attack path analysis and PDDL modeling, specifies the generation task, and provides constraints for producing valid attack paths.
Specifically, it consists of the following elements:
\begin{itemize}[leftmargin=1em]
    \item \emph{Role and Task}: Apart from defining the role, the prompt instructs the LLM to generate a syntactically correct, solvable, and semantically valid PDDL-encoded attack path from a given CVE description. When the description is incomplete or ambiguous, it instructs the LLM to make conservative and reasonable assumptions grounded in common security knowledge rather than introducing unsupported capabilities or attack steps.
    \item  \emph{Adversarial Model}: The prompt constrains the attacker's capabilities to ensure that generated attack path adheres to a realistic model. 
    The attacker's initial position and privilege level are consistent with the CVE's attack context. The attacker has no prior knowledge of or access to system internals, cannot break strong cryptography, and the attack exploits exactly one CVE to achieve exactly one STRIDE goal. The attacker initially holds no capabilities beyond what the authenticated sessions, elevated privileges, crafted artifacts, system internal secrets, and internal network sccess.
    The attack path may include user-interaction steps (e.g., clicking a link) when the CVE requires user interaction for exploitation.
    \item \emph{Attack Stage Reference}: The prompt provides a reference framework adapted for the Cyber Kill Chain~\cite{hutchins2011intelligence}, outlining the typical progression of a single-CVE exploitation through five stages: (i)~Exposure (mandatory), confirming the target is vulnerable (system technical composition, access paths, absent security controls, and other observable conditions); (ii) Weaponization, crafting or preparing the malicious artifact; (iii) Delivery, transmitting the malicious artifact or trigger to the target; (iv)~Exploitation, target-side processing that triggers the vulnerability, often decomposed into multiple actions; (v)~Impact (mandatory), achieving a STRIDE security outcome.
    This reference guides the LLM in structuring the attack path without enforcing a rigid linear progression, as not all vulnerabilities follow this exact sequence.
    \item  \emph{Instructions }: The prompt provides fine-grained instructions covering six quality dimensions aligned with the evaluation criteria defined in S3Eval~\cite{cui2026s3eval}: 
    \begin{itemize}[leftmargin=1em]
        \item \emph{Feasibility}: Each action must operate within the adversarial model and correspond to a realistic, executable attack operation.
        \item \emph{Atomicity}: Each action must represent a single, indivisible real-world attack step with justified effects.
        \item \emph{Completeness}: The attack path must be causally coherent with all critical intermediate steps present, progressing logically through attack stages.
        \item \emph{CVE Fidelity}: The path must correctly characterize the vulnerability, uses the right exploitation mechanism, and achieve an appropriate STRIDE goal.
        \item  \emph{Naming}: Action and predicate names must follow readable, descriptive, consistent conventions with clearly identifiable semantics and actors, e.g., attacker, target system, user, infrastructure. 
        \item \emph{PDDL Conformance}: Rules addressing common LLM errors in PDDL generation, including parameter scoping and predicate arity, identifier uniqueness, predicate block well-formedness, version comparison format, and parameter generalization.
        \end{itemize}
    \item  \emph{Output Format}: The prompt specifies the expected output structure. 
    A structured PDDL template can optionally be incorporated to provide explicit structural guidance. Details are described in \textbf{Output Template} below.
\end{itemize}
Listing~\ref{lst:system-prompt} and Listing~\ref{lst:output-template} respectively illustrate the simplified structure of the system prompt and the corresponding output template

\begin{promptbox}{Simplified System Prompt Structure.}
{lst:system-prompt}

\#\#\# \textbf{[ROLE]} \\
You are an expert in both attack path analysis and Planning Domain Definition Language (PDDL) modeling.

\vspace{0.8em}

\#\#\# \textbf{[TASK]} \\
You are given:\\
- A natural-language CVE vulnerability description

\vspace{0.3em}

Generate a PDDL-encoded attack path from the given CVE description. Output PDDL only, no explanation.

- Ensure the attack path is semantically valid, syntactically correct and solvable.

- When the CVE description is insufficient or ambiguous, make reasonable assumptions based on common security knowledge.

\vspace{0.8em}
\#\#\# \textbf{[ADVERSARIAL MODEL]} \\
The attacker's initial position and privilege level are consistent with the CVE's attack context. The attacker has no prior knowledge of or access to system internals, cannot break strong cryptography, and the attack exploits exactly one CVE to achieve exactly one STRIDE goal. ...

\vspace{0.3em}

The attacker initially holds no predicates from the following capability categories beyond what the CVE's attack context grants (illustrated with example predicate forms):

- credentials or tokens: (has-credentials), ...

...

- internal network access: (has-internal-network-access), ...

\vspace{0.8em}
\#\#\# \textbf{[ATTACK STAGE REFERENCE]} \\
Attack paths generally follow stages adapted from the Cyber Kill Chain, adapted to single-CVE exploitation:

1. Exposure (mandatory): ...

2. Weaponization: ...

3. Delivery: ...

4. Exploitation: ...

5. Impact (mandatory): ...

Each stage may require multiple atomic actions. Not all vulnerabilities follow this linear progression. ...

\vspace{0.8em}
\#\#\# \textbf{[INSTRUCTIONS]} \\
I1. Threat-Model Compliance. ...

I2. Real-World Executability. ...

...

I18. Version Variable Requirement. ...

\vspace{0.8em}
\#\#\# \textbf{[OUTPUT FORMAT]}\\
Output a complete PDDL domain in a Markdown code block (```pddl ...```). 

\vspace{0.3em}

Follow the same structure as the examples above. (few-shot only)

\vspace{0.3em}

Must include (:requirements :adl :fluents), ...

\vspace{0.3em}

Following the PDDL template below: <PDDL template> (optional)

\vspace{0.8em}
\#\#\# \textbf{[FEW-SHOT EXAMPLES]} (few-shot only)\\
<CVE description, reference PDDL-encoded attack path> pairs

\vspace{0.8em}
\#\#\# \textbf{[INPUT]} \\
CVE-XXXX-XXXXX

<CVE natural language description>

\end{promptbox}

    

\begin{promptbox}{Simplified Output Template Structure.}{lst:output-template}

{\ttfamily
(define (domain PDDL-AP-Domain)\\
\hspace*{1em}(:requirements :adl :fluents)\\
\hspace*{1em}(:functions (version ?Software) ...)\\
\hspace*{1em}(:types\\
\hspace*{2em}attacker user target-system infrastructure ...)\\
\hspace*{1em}(:constants CVE\_XXXX\_XXXXX - cve-identifier ...)\\
\hspace*{1em}(:predicates\\
\hspace*{2em}(spoofing ?Target - target-system)\\
\hspace*{2em}(tampering ?Target - target-system)\\
\hspace*{2em}(repudiation ?Target - target-system)\\
\hspace*{2em}(information-disclosure ?Target - target-system)\\
\hspace*{2em}(denial-of-service ?Target - target-system)\\
\hspace*{2em}(elevation-of-privilege ?Target - target-system) ...)
}

\vspace{0.8em}

{\ttfamily
;; Step 0: Vulnerability exposure (mandatory entry point)\\
(:action system-exposes-attack-surface-...\\
\hspace*{1em}:parameters (?Target - target-system ...)\\
\hspace*{1em}:precondition (and ...)\\
\hspace*{1em}:effect (and ...))
}

\vspace{0.8em}

{\ttfamily
;; Steps 1 to N-1: Intermediate attack actions\\
(:action actor-verb-description ...)
}

\vspace{0.8em}

{\ttfamily
;; Step N (final): Achieves exactly one STRIDE goal\\
(:action actor-achieves-impact\\
\hspace*{1em}:parameters (?Target - target-system ...)\\
\hspace*{1em}:precondition (and ...)\\
\hspace*{1em}:effect (and (denial-of-service ?Target)\\
\hspace*{1em};; or spoofing, tampering, repudiation, \\
\hspace*{1em};; information-disclosure,\\
\hspace*{1em};; elevation-of-privilege\\
\hspace*{1em}...))
}

\end{promptbox}
 
\textbf{Few-shot Learning.}
CVE2AP supports 0-shot, 1-shot, and 2-shot configurations. In the few-shot settings, examples are randomly sampled from a pool of expert-authored reference attack paths, while excluding the target CVE to prevent data leakage. Each example consists of a CVE description paired with its corresponding PDDL-encoded attack path and is presented as a use--assistent message pair. These examples provide the LLM with concrete demonstrations of the expected output structure, action granularity, style, and level of detail, thereby complementing the general constraints specified in the system prompt.

\textbf{Output Template.}
In the 0-shot setting, the prompt can optionally include a structured PDDL output template that provides an attack-path-specific domain skeleton. The template specifies the overall PDDL structure including \texttt{(:requirements)}, \texttt{(:types)}, \texttt{(:constants)}, \texttt{(:predicates)}, \texttt{(:functions)}, and action schema \texttt{(:action <action-name> :parameter(...), :precondition(...), :effects(...))}. 
It further guides the attack path progression from a mandatory vulnerability exposure action (Step~0) through intermediate exploitation steps to a final action achieving a single STRIDE impact (Step~N). 
This template guides the LLM toward syntactically conformant and domain-appropriate PDDL output without providing CVE-specific content.


\subsection{Error Feedback}
\label{sec:error-feedback}
A correct attack path should satisfy syntactic correctness, solvability, and semantic correctness~\cite{cui2026s3eval}.
Even with detailed instructions, due to limitations such as hallucinations, and factual inconsistencies~\cite{gu2026survey}, LLMs may still generate PDDL domains that contain errors, including invalid declarations or action configurations that do not admit a valid plan.
When generating PDDL-encoded attack paths,
representative errors include using variables in an action's \texttt{:precondition} or \texttt{:effect} that are not declared in the action's \texttt{:parameters} (parameter scoping violations), or writing negated predicates such as \texttt{(not (accepts-external-requests ?Sys))} in the \texttt{(:predicates)} construct, whose application is only permitted in action \texttt{:precondition} by PDDL syntax. 
To mitigate such errors, CVE2AP optionally incorporates an automated error feedback loop focusing on syntactic correctness and solvability, because existing domain-independent PDDL planners, such as Metric-FF planner~\cite{hoffmann2003metric}, can quickly and automatically detect these errors and provide concrete error information, such as syntax errors with their locations, which can be directly incorporated into iterative correction.
In contrast, although semantic correctness is essential, existing methods~\cite{cui2026s3eval} evaluating semantics primarily provide semantic quality judgments (i.e., a binary True/False judgment or a score ranging from 0 to 5 reflecting overall semantic quality) rather than feedback that identifies which elements of the generated attack path should be corrected.
We therefore leave semantic feedback as an important direction for future work.

Specifically, after the initial generation, a corresponding problem instance is automatically constructed from the domain.
Both the domain and problem files are then fed into Metric-FF planner to check syntax correctness and plan solvability.
Metric-FF detects syntax errors during its parsing phase and verifies whether a valid plan exists, with support for numeric fluents required for modeling version constraints in vulnerability exploitation. 
In our implementation, this checking process completes within milliseconds per domain, resulting in negligible overhead to the LLM-based generation process.
If the check fails, the specific error messages reported by the planner (e.g., undeclared predicates, type mismatches) are extracted and incorporated into a feedback prompt along with a concise set of common PDDL error correction rules.
This resulting feedback is appended to the existing conversation context as a new user message, thereby preserving the generated PDDL and the preceding reasoning context.
The LLM is then instructed to produce a revised PDDL domain based on the accumulated context. Listing~\ref{lst:feedback} shows the feedback message template.

This feedback loop allows up to three retry attempts within the same conversation, allowing the LLM to progressively correct errors while maintaining awareness of its prior outputs and the specific issues identified. If the generated PDDL passes the check before exhausting the retry budget, the process terminates early.  Otherwise, the procedure terminates after the third retry and returns the final generated output. 
The entire process is automated and does not require human intervention: the LLM receives only the planner-reported error signals and the predefined correction rules, enabling systematic correction while preserving the original generation context.


% ...

\begin{promptbox}{Error Feedback Message Template.}
{lst:feedback}

Metric-ff errors:\\
\texttt{\textless planner-reported error message\textgreater}

\vspace{0.5em}

Fix all errors in the domain. Critical rules:\\
(1) PARAMETER SCOPING: Every ?Var used in preconditions or effects Must appear in that action's :parameters.\\
(2) PDDL IDENTIFIER SYNTAX: Names must use only letters, digits, and hyphens. No dots allowed.\\
(3) DECLARED CONSTANTS ONLY: Every bare identifier (non-?var) must be declared in :constants.\\
\ldots\\
(6) PREDICATES BLOCK: Never write (not \ldots) inside :predicates declaration; (not \ldots) is only valid inside action :precondition.

\vspace{0.5em}

Output corrected PDDL only.

\end{promptbox}

\section{Evaluation}
\label{sec:evaluation}
 This section describes experimental setup (Section~\ref{sec:ex-setup}), the results (Sections~\ref{sec:ex-results}). 

\subsection{Experimental Setup}
Our objective is to systematically assess the effectiveness and cost of CVE2AP in generating PDDL-encoded attack paths, and to analyze the impact of key configuration decisions.
Accordingly, we address the following research questions:
\label{sec:ex-setup}
\begin{itemize} [leftmargin=3em, label=--]  
    \item[\textbf{RQ1:}] 
     \textbf{Effectiveness.} How effective is CVE2AP at automatically generating PDDL-encoded attack paths?
    \item[\textbf{RQ2:}]  
    \textbf{Cost.} What is the cost of CVE2AP for generating PDDL-encoded attack paths?
    \item[\textbf{RQ3:}]  
     \textbf{Configuration Impact.} How do different configuration decisions affect the effectiveness and cost of CVE2AP?
\end{itemize}

\noindent\textbf{Dataset.} 
We use 21 CVEs applied by S3Eval~\cite{cui2026s3eval}, covering diverse vulnerability types. Each CVE includes a natural language description and expert-authored reference PDDL-encoded attack path.

\noindent\textbf{Generation LLM models.}
We select five LLM models spanning both online commercial and locally-hosted options at different scales to examine the trade-off between capability and deployment cost.
\begin{itemize}[leftmargin = 1em]
    \item \emph{Online commercial:} \texttt{GPT-5.5}~\cite{openai_gpt55_2026} (latest flagship), and \texttt{Deepseek (DS) V4-PRO}~\cite{guo2025deepseek}. 
    \item  \emph{Locally-hosted:} \texttt{Qwen3-\{4B, 8B, 14B, 32B\}}~\cite{zhang2025qwen3}, which share the same architecture at different scales, enabling analysis of how model capacity affects judgment quality in a cost-effective, self-hosted setting. 
\end{itemize} 

\noindent \textbf{Configuration Space.} 
Each generation run is defined by three configuration dimensions:
\begin{itemize}[leftmargin = 1em]
    \item \emph{Few-shot learning:} 0-shot, 1-shot, and 2-shot. Few-shot examples are randomly selected from the reference pool, excluding the target CVE.
    \item \emph{Error feedback:} After the initial generation (which proceeds without prior indication of feedback mechanism in the prompt), if the output fails syntax or solvability checks, the specific error messages are fed back to the LLM for up to 3 retry attempts within the same conversation context.
    \item \emph{Output template:} For different shot mode, a structured PDDL output template can be provided to guide the generation format. We evaluate both with and without the output template.
\end{itemize}
In total, the above three dimensions yield $3 \times 2 \times 2 =12$ prompt configurations per model (\{0-shot, 1-shot, 2-shot\} $\times$ \{with, without error feedback\} $\times$ \{with, without output template\}).

\noindent \textbf{Evaluation Metrics.} 
We evaluate the generated PDDL-encoded attack path using the S3Eval pipeline~\cite{cui2026s3eval}, which comprehensively assess the quality of PDDL-encoded AP through a 3-layer structure:
\begin{itemize}[leftmargin = 1em]
    \item \emph{S1 (Syntax):} Whether the generated PDDL-encoded APs conforms to the PDDL specification, including valid structure, consistent type declaration, correct predicate arities, and well-defined action schemas.
    \item  \emph{S2 (Solvability):} Whether the domain, together with an automatically generated problem file, admits at least one valid plan, verified by the Metric-ff~\cite{hoffmann2003metric}.
    \item \emph{S3 (Semantic):} Whether the domain faithfully models the intended attack path. Semantic quality is assessed through embedding-based similarity (\texttt{Qwen3-Embedding-8B}~\cite{zhang2025qwen3}) and LLM-as-expert (\texttt{Deepseek-R1}~\cite{guo2025deepseek}), each operating in both \emph{intrinsic} (candidate vs. CVE natural language description) and \emph{extrinsic} (candidate vs. expert-authored reference) settings. The LLM-as-expert applies two rubric variants: \emph{simplified}, which uses concise criterion descriptions to produce a holistic binary judgment (True/False), and  \emph{detailed}, which provides fine-grained instructions with violation examples and scores each criterion on a 0--5 scale (0--2 = violation, 3--5 = compliance), with the minimum score across all criteria deriving a binary signal. 
\end{itemize}
We report the \textbf{\emph{correct rate}} (proportion of the generations that assessed as correct at each layer) as the primary metric, and the average \textbf{\emph{generation time}} and \textbf{\emph{token consumption}} per sample as cost metrics.

\noindent\textbf{Testbed.}
We conduct all experiments using Python 3.10 on two machines: 
a macOS laptop equipped with an Intel Core i7-8559U CPU and Intel Iris Plus Graphics 655 (16\,GB RAM),
and a Linux server equipped with an NVIDIA RTX~4090 GPU (24\,GB VRAM, CUDA~13.2) and 62\,GB RAM. L
Online LLM judges are assessed via API with temperature set to 0, and prompted via chat- or completion-based interfaces depending on model availability. Local models are served via \texttt{llama.cpp}~\cite{llamacpp} with quantized weights (Q8\_0 for \texttt{Qwen3-\{4B, 8B, 14B\}}, Q4\_K\_M for \texttt{Qwen3-32B}).
We use a fixed random seed of 42 for reproducibility. 
We note that \texttt{GPT-5.5} is subject to OpenAI's content safety filter (\texttt{cyber\_policy}), which non-deterministically blocks a negligible fraction of the generation of PDDL-encoded attack paths. 
When this occurs, we retry the generation request until a valid generation is obtained. This issue can also be solved by applying for OpenAI's Research Access Program, which grants exemptions from certain content filters.

\subsection{Results}
\label{sec:ex-results}
\subsubsection{RQ1 (Effectiveness)}
\begin{figure*}[!h]
    \centering
    \includegraphics[width=0.9\linewidth]{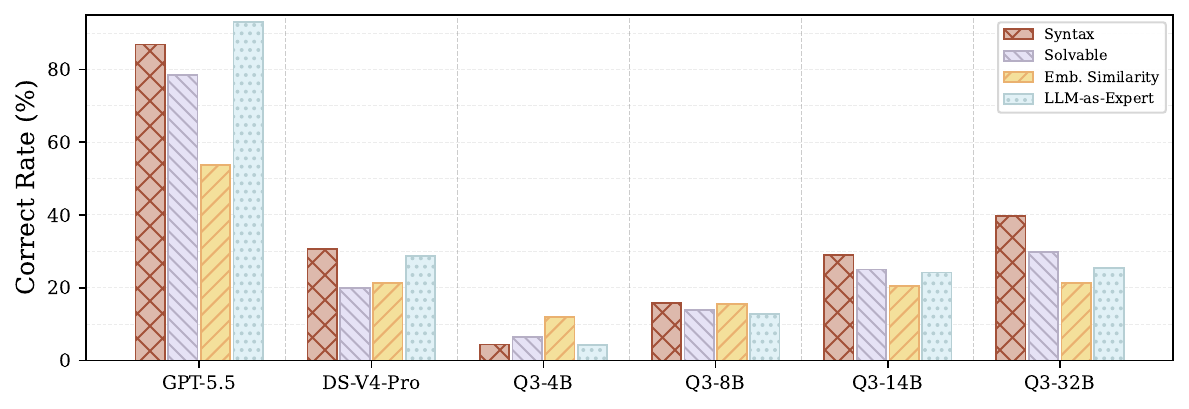}
    \caption{Overall generation effectiveness across models.}
    \label{fig:rq1-effectiveness}
\end{figure*}

Figure~\ref{fig:rq1-effectiveness} reports the correct rate for each evaluation dimension across syntax, solvability, and semantic quality (embedding-based similarity and LLM-as-expert) of all six LLM models. Each bar indicates the mean correct rate over the 12 prompt configurations.

Overall, CVE2AP is effective in automatically generating high-quality PDDL-encoded attack paths. With the best-performing model, \texttt{GPT-5.5}, CVE2AP achieves a correct rate of 86.9\% for syntax, 78.6\% for solvability, 53.8\% for embedding-based similarity, and 93.1\% for LLM-as-expert. 
Additionally, the performance of locally-hosted models improves consistently with model size across all evaluation dimensions
The weaker performance of smaller local models can be attributed to: (i)~limited capability for the multi-constraint reasoning to generate PDDL-encoded attack paths demands; (ii)~limited exposure during pre-training to PDDL planning language and cybersecurity, particularly CVE domain knowledge; and (iii)~context window limitations. Specifically, few-shot examples, error feedback, the output template increase the input context, which may overwhelm smaller models, such as \texttt{Qwen3-4B} and \texttt{Qwen3-8B}, resulting in empty, ineffective, or malformed outputs.
Notably, despite these challenges, \texttt{Qwen3-32B} achieves a syntax correct rate of 39.7\% , produces solvable attack paths for approximately one-third of its generations, and satisfies semantic requirements in more than 20\% of cases under both embedding-based similarity and LLM-as-expert evaluation.
These results are encouraging given the inherent complexity of generating PDDL-encoded attack paths from a CVE description, which demands simultaneously satisfying syntax correctness, solvability, and semantic requirements including action feasibility, action atomicity, and attack stage completeness.

 \begin{tcolorbox}[ 
  colback=gray!10,
  colframe=black,
  boxrule=0.2mm,
  arc=1mm,
  left=1mm,
  right=1mm,
  top=1mm,
  bottom=1mm,  
    pad at break*=0mm,   
]
\textbf{RQ1 Summary: }
CVE2AP effectively generates PDDL-encoded attack paths that satisfy syntactic correctness (up to 86.9\% of generations), solvability (up to 78.6\%), and semantic quality (53.8\% in embedding-based similarity and 93.1\% in LLM-as-expert evaluation)
\end{tcolorbox}

\subsubsection{RQ2 (Cost)}
\begin{figure*}[!h]
    \centering
    \begin{subfigure}{0.48\linewidth}
        \includegraphics[width=\linewidth]{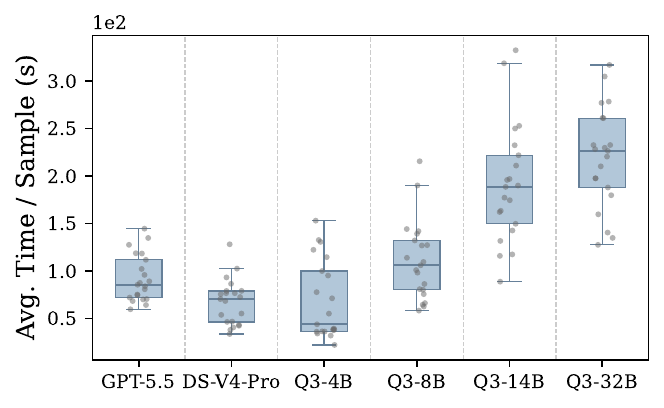}
        \caption{Avg. Time / Sample}
        \label{fig:rq2-time}
    \end{subfigure}
    \hfill
    \begin{subfigure}{0.48\linewidth}
        \includegraphics[width=\linewidth]{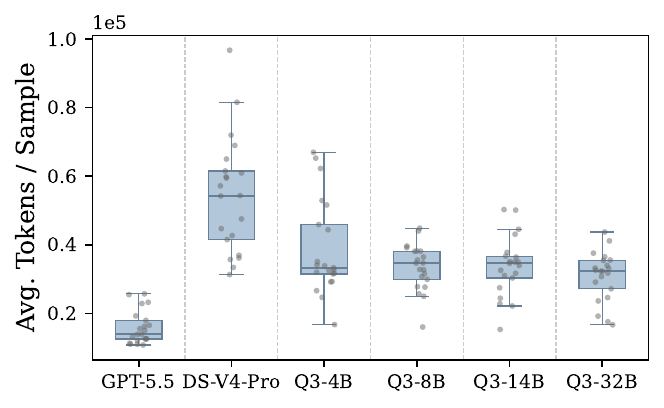}
        \caption{Avg. Tokens / Sample}
        \label{fig:rq2-tokens}
    \end{subfigure}
    \caption{Per-sample generation cost across models.}
    \label{fig:rq2-cost}
\end{figure*}

Figure~\ref{fig:rq2-cost} (a) and (b) show the per-sample generation time and token cost, respectively, with box plots showing Q1-median-Q3. Each dot represents the mean cost of generating attack paths for one CVE across all 12 configurations.

For generation time, we make two observations: (i)~the per-sample generation time increases monotonically with model size, with the median increasing from approximately 40\,s for \texttt{Qwen3-4B} to 215\,s for \texttt{Qwen3-32B}, as larger local models require longer time to inference; (ii)~ the commercial models generate faster on average despite their larger model sizes, with median generation time of approximately 85\,s for \texttt{GPT-5.5}, 65\,s for \texttt{Deepseek-V4-Pro}. This situation benefits from their optimized serving infrastructures.

For token cost, we have three observations: (i)~\texttt{Qwen3-4B} has the lowest token consumption, consuming approximately 10,000 tokens per sample on average, substantially fewer than all other models; (ii)~\texttt{Deepseek-V4-Pro} consumes the most (about 53,000 tokens per-generation), which may attributed to its additional reasoning process and more verbose output style; (iii)~the token cost of local \texttt{Qwen3} models remains relatively stable across difference model scales, with the median value ranging from about 32,000 to 34,000.

Overall, \texttt{GPT-5.5} achieves best quality-cost tradeoff: it outputs highest-quality PDDL-encoded attack paths, while being the fastest commercial model and maintaining the lowest token cost.

 \begin{tcolorbox}[ 
 breakable,
  colback=gray!10,
  colframe=black,
  boxrule=0.2mm,
  arc=1mm,
  left=1mm,
  right=1mm,
  top=1mm,
  bottom=1mm,
    pad at break*=0mm,   
]
\textbf{RQ2 Summary:} 
CVE2AP achieves a favorable quality-cost tradeoff, with \texttt{GPT-5.5} providing the highest generation quality, generating fastest among commercial models, and consuming the lowest tokens. 
\end{tcolorbox}

\subsubsection{RQ3 (Configuration Impact)}
\begin{figure*}[!h]
    \centering
    \begin{subfigure}{0.28\linewidth}
        \includegraphics[width=\linewidth]{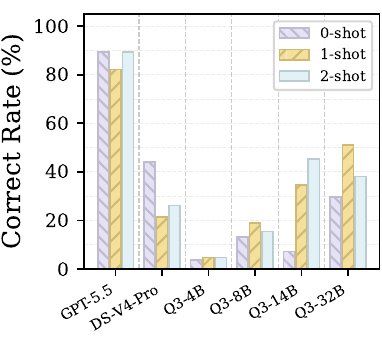}
        \caption{S1 Syntax}
        \label{fig:rq3-fs-s1}
    \end{subfigure}
    \hfill
    \begin{subfigure}{0.23\linewidth}
        \includegraphics[width=\linewidth]{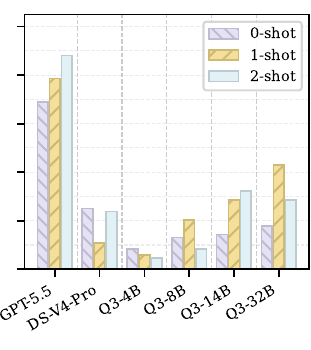}
        \caption{S2 Solvability}
        \label{fig:rq3-fs-s2}
    \end{subfigure}
    \hfill
    \begin{subfigure}{0.23\linewidth}
        \includegraphics[width=\linewidth]{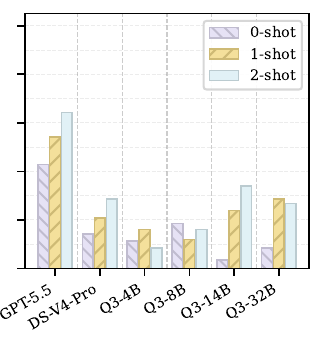}
        \caption{S3 Emb. Similarity}
        \label{fig:rq3-fs-s3emb}
    \end{subfigure}
    \hfill
    \begin{subfigure}{0.23\linewidth}
        \includegraphics[width=\linewidth]{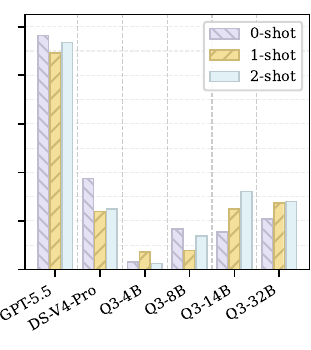}
        \caption{S3 LLM-as-Expert}
        \label{fig:rq3-fs-s3llm}
    \end{subfigure}
    \caption{Impact of few-shot count on generation quality. }
    \label{fig:rq3-fewshot}
\end{figure*}

\begin{figure*}[!h]
    \centering
    \begin{subfigure}{0.48\linewidth}
        \includegraphics[width=\linewidth]{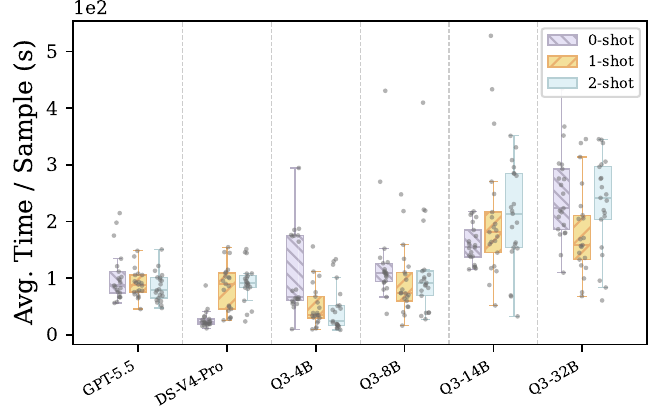}
        \caption{Avg. Time / Sample}
        \label{fig:rq3-fs-cost-time}
    \end{subfigure}
    \hfill
    \begin{subfigure}{0.48\linewidth}
        \includegraphics[width=\linewidth]{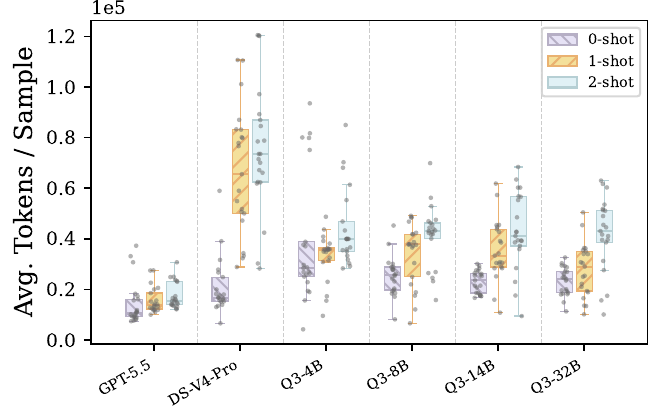}
        \caption{Avg. Tokens / Sample}
        \label{fig:rq3-fs-cost-tokens}
    \end{subfigure}
    \caption{Impact of few-shot count on generation cost.}
    \label{fig:rq3-fewshot-cost}
\end{figure*}

\begin{figure*}[!h]
    \centering
    \begin{subfigure}{0.28\linewidth}
        \includegraphics[width=\linewidth]{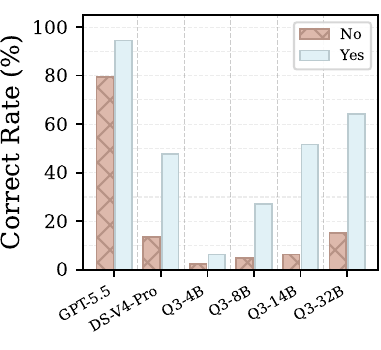}
        \caption{S1 Syntax}
        \label{fig:rq3-fb-s1}
    \end{subfigure}
    \hfill
    \begin{subfigure}{0.23\linewidth}
        \includegraphics[width=\linewidth]{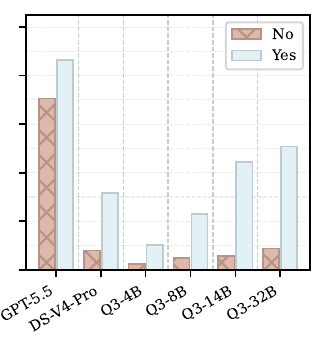}
        \caption{S2 Solvability}
        \label{fig:rq3-fb-s2}
    \end{subfigure}
    \hfill
    \begin{subfigure}{0.23\linewidth}
        \includegraphics[width=\linewidth]{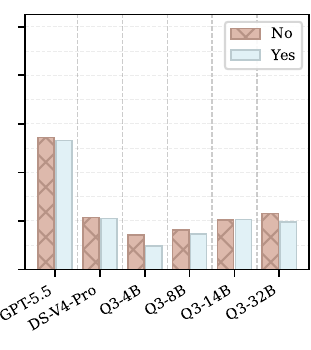}
        \caption{S3 Emb. Similarity}
        \label{fig:rq3-fb-s3emb}
    \end{subfigure}
    \hfill
    \begin{subfigure}{0.23\linewidth}
        \includegraphics[width=\linewidth]{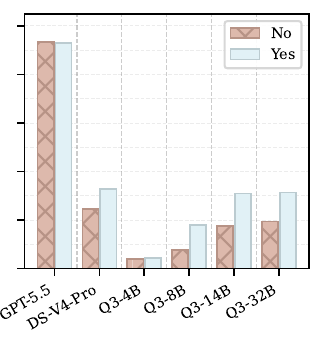}
        \caption{S3 LLM-as-Expert}
        \label{fig:rq3-fb-s3llm}
    \end{subfigure}
    \caption{Impact of error feedback on generation quality. }
    \label{fig:rq3-feedback}
\end{figure*}

\begin{figure*}[!h]
    \centering
    \begin{subfigure}{0.48\linewidth}
        \includegraphics[width=\linewidth]{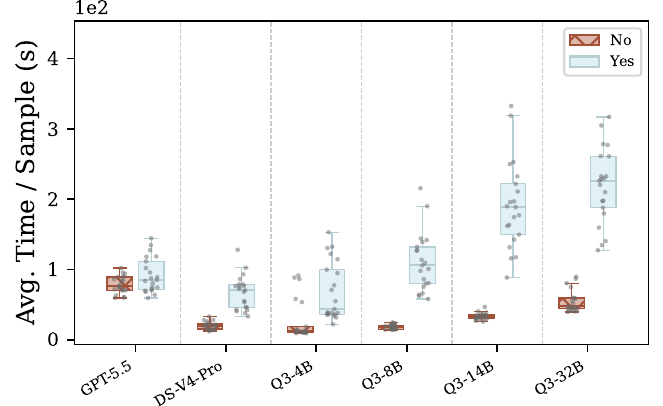}
        \caption{Avg. Time / Sample}
        \label{fig:rq3-fb-cost-time}
    \end{subfigure}
    \hfill
    \begin{subfigure}{0.48\linewidth}
        \includegraphics[width=\linewidth]{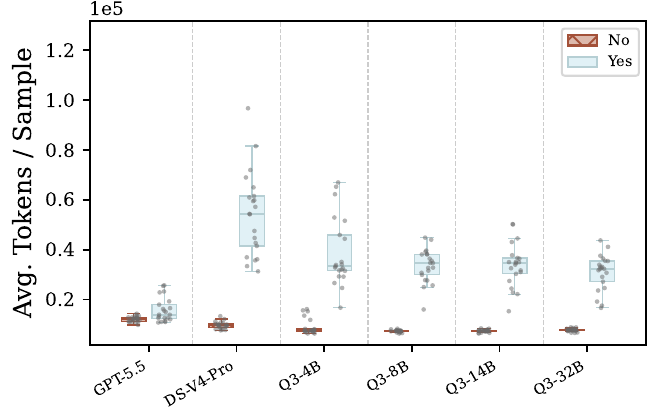}
        \caption{Avg. Tokens / Sample}
        \label{fig:rq3-fb-cost-tokens}
    \end{subfigure}
    \caption{Impact of error feedback on generation cost. }
    \label{fig:rq3-feedback-cost}
\end{figure*}

\begin{figure*}[!h]
    \centering
    \begin{subfigure}{0.28\linewidth}
        \includegraphics[width=\linewidth]{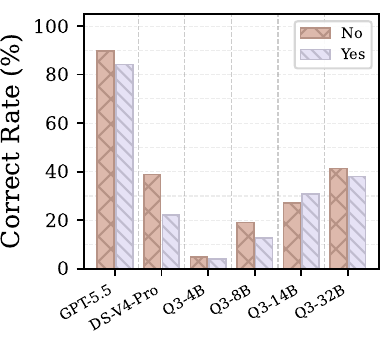}
        \caption{S1 Syntax}
        \label{fig:rq3-tmpl-s1}
    \end{subfigure}
    \hfill
    \begin{subfigure}{0.23\linewidth}
        \includegraphics[width=\linewidth]{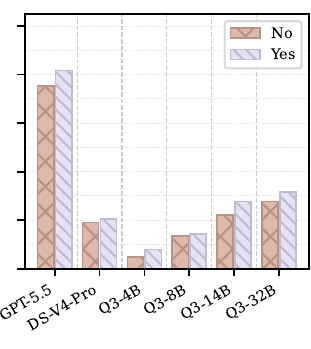}
        \caption{S2 Solvability}
        \label{fig:rq3-tmpl-s2}
    \end{subfigure}
    \hfill
    \begin{subfigure}{0.23\linewidth}
        \includegraphics[width=\linewidth]{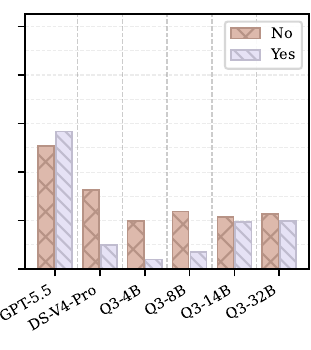}
        \caption{S3 Emb. Similarity}
        \label{fig:rq3-tmpl-s3emb}
    \end{subfigure}
    \hfill
    \begin{subfigure}{0.23\linewidth}
        \includegraphics[width=\linewidth]{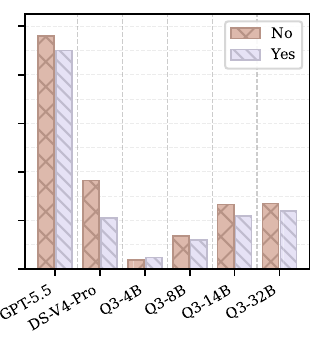}
        \caption{S3 LLM-as-Expert}
        \label{fig:rq3-tmpl-s3llm}
    \end{subfigure}
    \caption{Impact of output template on generation quality. }
    \label{fig:rq3-template}
\end{figure*}

\begin{figure*}[!h]
    \centering
    \begin{subfigure}{0.47\linewidth}
        \includegraphics[width=\linewidth]{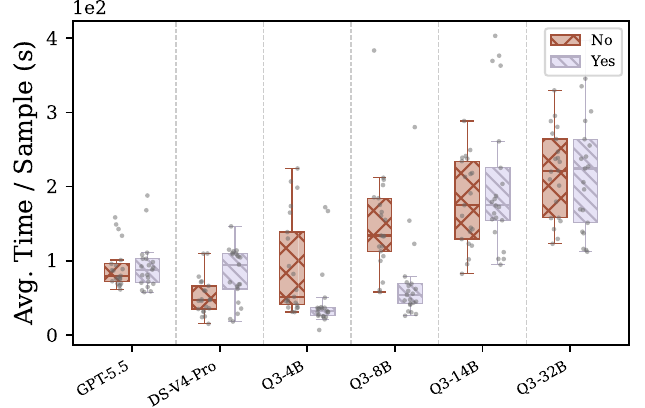}
        \caption{Avg. Time / Sample}
        \label{fig:rq3-tmpl-cost-time}
    \end{subfigure}
    \hfill
    \begin{subfigure}{0.48\linewidth}
        \includegraphics[width=\linewidth]{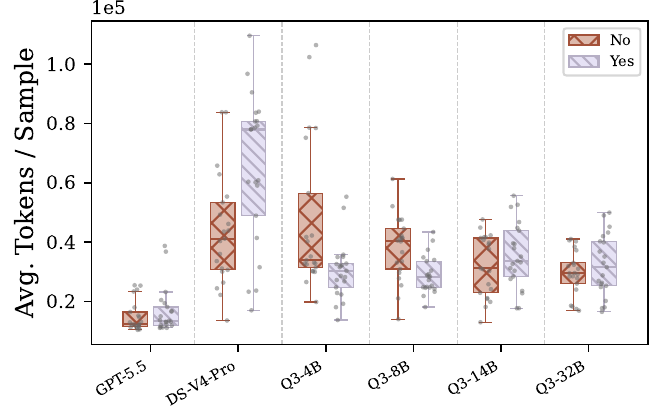}
        \caption{Avg. Tokens / Sample}
        \label{fig:rq3-tmpl-cost-tokens}
    \end{subfigure}
    \caption{Impact of output template on generation cost.}
    \label{fig:rq3-template-cost}
\end{figure*}

We investigate three configuration dimensions: the number of few-shot examples, error feedback, and the output template. For each dimension, we evaluate its impact on both generation quality (Figures~\ref{fig:rq3-fewshot}--\ref{fig:rq3-template}) and generation cost (Figures~\ref{fig:rq3-fewshot-cost}--\ref{fig:rq3-template-cost}).

\noindent\textbf{Few-shot examples.}
Figure~\ref{fig:rq3-fewshot} shows the impact of the number of few-shot examples (0, 1, and 2) on generation quality. Overall, the effect varies across LLM models. 
For \texttt{GPT-5.5}, the correct rates of solvability and embedding-based similarity increase consistently with the number of few-shot examples. 
In contrast, the correct rates of syntax and LLM-as-expert are comparable with 0-shot or 2-shot prompting, suggesting that \texttt{GPT-5.5} derives limited additional benefit from exemplars, as it contains internal knowledge to support high-quality attack-path generation. 
For the local models, increasing the number of few-shot examples stably benefits \texttt{Qwen3-14B}, whereas \texttt{\{Qwen3-4B, 8B\}} show no consistent trend.
Although exemplars provide useful structural guidance that can partially compensates for internal limited PDDL knowledge, the additional examples substantially extend the input context. Thus, for the smaller models (\texttt{\{Qwen3-4B, 8B\}}), this increased context my exceed their effective context-processing capacity, resulting in empty, ineffective, or malformed outputs.

Figure~\ref{fig:rq3-fewshot-cost} shows the cost impact. 
In general, both generation time and token cost increase monotonically with the number of few-shot examples across the evaluated models, as additional exemplars increase the amount of context processed during generation. 
The increase is particularly significant for \texttt{DS-V4-Pro} in terms of token consumption, with the 1-shot and 2-shot settings requiring more than twice as many tokens as the 0-shot setting. 

Overall, additional few-shot examples do not reliably improve the generation quality across the evaluated models, while consistently increasing the generation cost.

\noindent\textbf{Error feedback.}
Figure~\ref{fig:rq3-feedback} compares generation quality with (`Yes') and without (`No') error feedback. 
Unlike few-shot examples, whose effects vary substantially across models and evaluation dimensions, error feedback consistently improves generation quality in syntax (up to +49 percentage points), solvability (up to +41 percentage points), and LLM-as-expert semantic evaluation (up to +12 percentage points). 
In contrast, a slight degradation is observed in embedding-based semantic similarity. 
These results suggest that error feedback helps LLMs correct syntactic and solvability errors in subsequent retries. The limited improvement in embedding-based semantic quality is expected because the current feedback mechanism targets syntax and solvability errors rather than semantic discrepancies. Future work could incorporate semantic errors into the feedback loop to further improve embedding-based semantic similarity.

Figure~\ref{fig:rq3-feedback-cost} reports the corresponding cost impact. 
Obviously, error feedback consistently increase the per-sample generation time cost and generation token consumptions, because feedback error message for failed attempt triggers additional retry and require the model to process the extended context. 

Overall, error feedback provides a favorable quality-cost trade-off, substantially improving generation quality at a relatively modest additional cost.

\noindent\textbf{Structured output template.}
Figure~\ref{fig:rq3-template} compares the generation quality with (`Yes') and without (`No') the output template. 
The impact of the output template varies across models and evaluation dimensions.
Overall, the template provides limited benefits, improve the correct rate of solvability by approximately 1--5 percentage points across models. The improvement can be attributed to that the template guides LLMs toward outputs that conform to the required PDDL structure. However, for the weaker local models in the \texttt{Qwen3} series, the additional instructions introduced by the template increase prompt complexity and may make it more difficult to follow the required constraints, resulting in more incorrect outputs in syntax and semantic evaluation. 

Figure~\ref{fig:rq3-template-cost} shows that the output template generally increases per-sample generation time and token consumption, with some variations across models. Overall, the output template provides limited and model-dependent quality benefits while incurring additional generation costs.


 \begin{tcolorbox}[ 
 breakable,
  colback=gray!10,
  colframe=black,
  boxrule=0.2mm,
  arc=1mm,
  left=1mm,
  right=1mm,
  top=1mm,
  bottom=1mm,
  pad at break*=0mm,   
]
\textbf{RQ3 Summary:}  
Error feedback is the most consistently beneficial configuration,
while the effects of few-shot examples and output templates vary across models.
\end{tcolorbox}

\section{Related Work}
\label{sec:related-work}

\subsection{PDDL-based Attack Path Modeling} \label{sec:rl-pddlap}
AI planning provides a formal framework for modeling multi-step cyber attacks by representing attack actions as planning operators with explicit preconditions and effects. Among existing planning formalisms, the PDDL has been widely adopted for this purpose. Roberts et al.~\cite{roberts2011personalized} demonstrated the use of PDDL for personalized vulnerability analysis by encoding attack actions as PDDL schemas and incorporating user activities and system-specific configurations to automatically identify paths targeting to a specific system.
Falco et al.~\cite{falco2018master} formulated adversarial strategy generation for critical infrastructures as an AI planning problem. 
More recently, Bashir et al.~\cite{bashir2024resiliency} integrated PDDL-based attack graphs with fault trees to support resilience analysis of cyber-physical systems, while Inayat et al.~\cite{inayat2025employing} modeled cyber attacks in PDDL to generate attack steps,  
and recommend security controls.

As attack scenario evolve increasingly complex, prior work has also investigated the scalability and maintainability of PDDL models. PDDLAssistant~\cite{tiwary2017pddlassistant} supports the incremental development of PDDL-encoded attack graphs by reconciling different versions of PDDL representations and explaining changes between generated plans.
Wang et al.~\cite{wang2021automatic} improved the efficiency of PDDL-based attack planning in large-scale networks through a device reachability graph partitioning approach. 
Despite these efforts, most existing approaches assume that PDDL attack models are manually constructed. 
In practice, this construction process is labor-intensive and requires expertise in both cyber security and AI planning to specify predicates, action schemas, and their corresponding preconditions and effects. This modeling burden becomes particularly significant as vulnerabilities, attack techniques, system and environmental conditions continuously evolve.

To reduce this modeling effort, Bezawada et al. ~\cite{bezawada2019agbuilder} proposed to automatically derive PDDL attack paths from CVE description using traditional NLP techniques, including Part-Of-Speech (POS) tagging and rule-based extraction. However, the reliance on handcrafted extraction rules limits the approach's ability to capture the diverse linguistic expressions and incomplete, implicit and heterogeneous information commonly contained in vulnerability descriptions and threat intelligence. Furthermore, the approach was evaluated only on a limited case study without systematic evaluation across multiple important quality metrics, such as syntax correctness, solvability, semantic correctness.

Overall, existing research has established PDDL as an effective formalism for representing multi-step cyber attacks, while automatic construction of high-quality, planner-ready PDDL attack paths remains unexplored. 
In particular, generating attack-specific PDDL representations from unstructured vulnerability descriptions remains challenging because the generated actions must satisfy not only the formal planning constraints and cybersecurity-specific semantic requirements, such as faithful representation of vulnerability exploitation mechanisms, and logically complete attack progression.
Our work addresses this gap by leveraging LLMs to automate the construction of PDDL-encoded attack paths from CVE descriptions and systematically evaluating their syntactic correctness, solvability, and semantic correctness.

\subsection{LLM-based PDDL Domain Generation} \label{rl-llmppddl}
Recent advances in LLMs have enabled automatic generation of PDDL planning domains from natural language descriptions. 
Guan et al.~\cite{guan2023leveraging} investigated the use of pre-trained LLMs for PDDL domain construction, demonstrating the potential of LLMs to translate natural-language descriptions into formal planning representations.
Smirnov et al. ~\cite{smirnov2024generating} further improved the generation quality by incorporating automatic consistency checking and reachability analysis into an iterative refinement loop.
Oswald et al.~\cite{oswald2024large} and Huang et al.~\cite{huang2025limit} systematically evaluated the ability of LLMs to generate PDDL domains from natural-language descriptions, highlighting both their potential and remaining challenges in producing semantically correct and executable planning domains. 
More recently, Huang et al.~\cite{huang2025planning} proposed a fully automated pipeline that leverages multiple LLMs to convert natural-language, step-wise action descriptions into candidate PDDL specifications. These candidates are ranked by embedding-based semantic similarity and further filtered through planner-based solvability checking to identify suitable PDDL domains.

Despite these advances, existing LLM-based approaches primarily target general-purpose planning domains and provide limited consideration of cybersecurity-specific attack-path modeling.
Unlike general planning domains, attack path modeling introduces additional domain-specific semantic requirements beyond syntax correctness and planning solvability. Generated attack actions must, for example, accurately reflect the vulnerability's exploitation mechanism, and preserve the causal dependencies required for a complete attack progression. 
These requirements introduce a fundamental distinction between a PDDL domain that is formally valid and solvable and one that faithfully represents a real-world cyber attack.


Our work investigates the application of LLMs to automatically generate PDDL-encoded attack paths from CVE descriptions.
To assess the resulting attack paths beyond formal PDDL correctness, we employ S3Eval~\cite{cui2026s3eval} to evaluate three complementary dimensions: syntactic correctness, solvability, and semantic correctness.  This allows us to investigate whether LLM-generated PDDL representations are not only planner-executable but also faithful to the underlying vulnerability-exploiting-specific semantics

\section{Conclusion}
\label{sec:conclusion}
The rapid evolution of the cyber threat landscape has rendered the manual construction and maintenance of PDDL-encoded attack paths impractical, making automated attack path generation imperative. 
This paper introduced \textbf{CVE2AP}, an LLM-based approach for automatic generation of PDDL-encoded attack paths from natural language CVE descriptions.
By combining specifically designed structured prompting with an iterative, domain-independent planner-guided error-feedback mechanism, CVE2AP improves the syntactic correctness and planning solvability of generated attack paths while preserving their semantic quality.
Our systematic evaluation applying S3Eval~\cite{cui2026s3eval} framework across diverse LLMs and generation configurations demonstrates that CVE2AP is capable of generating high-quality PDDL-encoded attack paths 
with reasonable generation time and token consumption. 
The results further reveal substantial differences in the quality-cost profiles of the evaluated LLMs, with \texttt{GPT-5.5} providing the best generation performance and the most favorable quality-cost trade-off. Furthermore, configuration analysis shows that error feedback is the most consistently effective strategy for improving generation quality, whereas the benefits of few-shot examples and structured output templates diverse.
These findings highlight the importance of jointly considering model capability, generation cost, and configuration strategy when deploying LLM-based methodology to generate PDDL-encoded attack paths.
Despite these encouraging results, the current implementation of CVE2AP focuses on attack paths involving the exploitation of a single vulnerability. As future work, we plan to extend the approach to support the generation of multi-vulnerability attack paths and chained exploitations.

\section{Data Availability}
\label{sec:data-availability}

The replication package, including the source code, experimental data, and scripts required to reproduce our experiments, is publicly available at \url{https://github.com/LincuiDudu/pddl-ap-generation}.


%


\balance

\bibliographystyle{IEEEtran}
\bibliography{IEEEabrv,references}

\end{document}